\documentclass[letterpaper]{article}
\usepackage{aaai2027}
\usepackage{graphicx}
\usepackage{booktabs}
\usepackage{multirow}
\usepackage{natbib}
\usepackage{url}
\usepackage{amsmath}
\usepackage{amssymb}
\usepackage{cuted}
\usepackage{capt-of}
\usepackage{booktabs}
\usepackage{tabularx}
\usepackage{array}
\usepackage{url}
\title{PEAK: Precise and Persistent Concept Erasure via k-Sparse Autoencoders}

\author{
    Man Jiang$^{1,*}$,
    Ouxiang Li$^{2,*}$,
    Weibao Xue$^{1}$,
    Zhenhua Tang$^{3}$,\\
    Yuan Wang$^{2}$,
    Shuo Wang$^{2}$,
    Yanbin Hao$^{1,\dagger}$
}

\affiliations{
    $^{1}$Hefei University of Technology, 
    $^{2}$University of Science and Technology of China, 
    $^{3}$University of Macau \\
    \texttt{jiangman1224@gmail.com, haoyanbin@hfut.edu.cn}
}

\begin{document}
\setcounter{secnumdepth}{2}
\maketitle
\renewcommand{\thefootnote}{\fnsymbol{footnote}}
\footnotetext[1]{Equal Contributions.}
\footnotetext[2]{Corresponding author.}
\renewcommand{\thefootnote}{\arabic{footnote}}

\begin{figure*}[ht]
\centering
\includegraphics[width=\textwidth,]{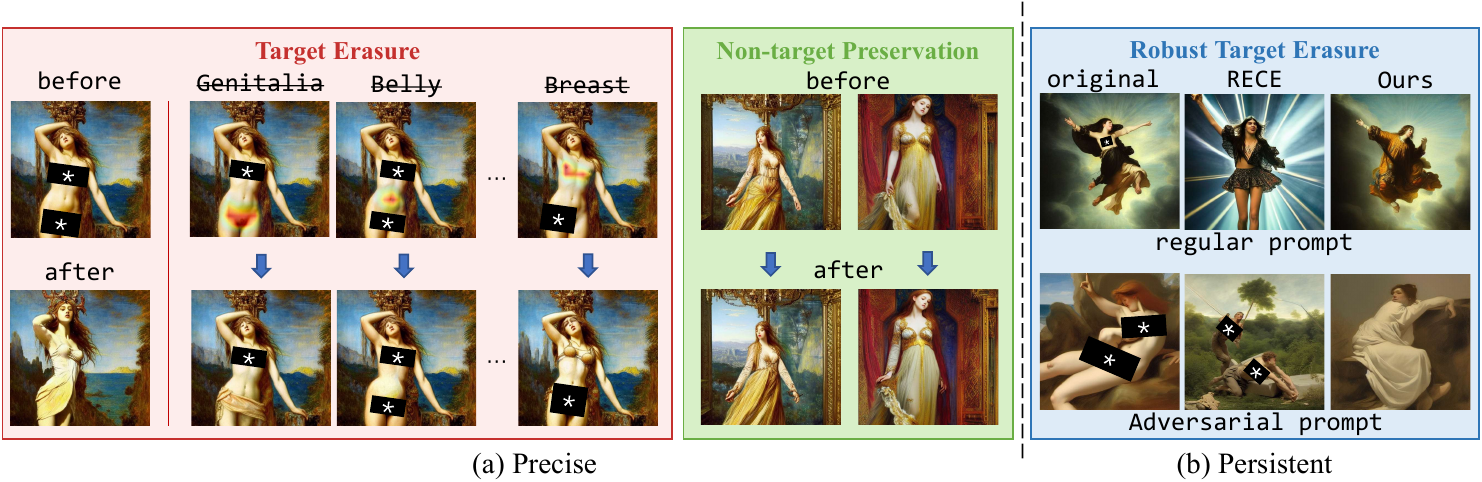}
\caption{
Two key characteristics of PEAK for concept erasure in diffusion models.
(a) Precise: PEAK removes the target concept (\textit{i.e.}, the concepts should be erased) while preserving non-target concepts (\textit{i.e.}, the concepts should be retained).
(b) Persistent: PEAK remains effective under both regular and adversarial prompts.
}
\label{fig:figure1}
\end{figure*}
\begin{abstract}
Erasing concepts from large-scale text-to-image (T2I) diffusion models has become increasingly crucial due to the growing concerns over copyright infringement, privacy violations, and offensive content. Existing approaches struggle to achieve both precise and persistent concept erasure: inaccurate localization of concept-related representations may cause unintended semantic interference, while incomplete removal of the underlying concept knowledge allows adversarial recovery. To address this dilemma, we propose PEAK, a \textbf{\textit{precise}} and \textbf{\textit{persistent}} concept erasure framework via k-Sparse Autoencoders (kSAEs). PEAK first trains a kSAE on internal activations of the diffusion denoising network to decompose dense representations into interpretable sparse features. By contrasting sparse activations induced by target and non-target prompts, PEAK identifies a compact set of target-specific features according to both activation strength and frequency. These localized features are then used for parameter optimization, where PEAK selectively suppresses target-related activations while preserving complementary non-target ones towards the original model. This feature-guided optimization embeds concept erasure directly into diffusion parameters, eliminating the need for additional inference-time intervention and facilitating effective persistence against adversarial attacks. Extensive experiments demonstrate that PEAK achieves effective and robust concept erasure. On the I2P benchmark, PEAK reduces NudeNet detections from 582 to 6, lowers the average attack success rate (ASR) from 96.52\% to 5.63\%, and preserves general generation quality on MS-COCO with a near-zero KID.
Our code and models are available at: \url{https://github.com/manmanTAT/PEAK}
\end{abstract}
\maketitle

\section{Introduction}
Text-to-image (T2I) models can easily generate high-quality images from natural language prompts~\cite{ho2020denoising}, but they may also produce copyrighted, pornographic, or privacy-sensitive content inherited from large-scale web data~\cite{shan2023glaze,schramowski2023safe,carlini2023extracting,li2026easier}. Concept erasure aims to remove such unwanted concepts from a pretrained model without retraining it from scratch~\cite{gandikota2023erasing}. Despite the diverse strategies developed for concept erasure, they all depend on a more fundamental question: \textbf{\textit{what exactly should be erased inside the model?}} Without accurately localizing the internal features responsible for generating the target concept, concept erasure suffers from both incomplete removal and unintended semantic damage: residual target information may remain recoverable under adversarial prompts, while modifications to entangled non-target features can degrade unrelated semantics and overall generation quality. Therefore, concept erasure should be both \textbf{\textit{precise}}, removing the target concept while preserving unrelated semantics, and \textbf{\textit{persistent}}, preventing the erased concept from being recovered by adversarial concept-recovery attacks~\cite{zhang2024defensive}. At its core, effective concept erasure hinges on accurate feature localization and selective suppression.

In this context, k-sparse autoencoders (kSAEs)~\cite{makhzani2014ksparseautoencoders} provide a natural basis for concept erasure by decomposing dense neural activations into sparse, semantically meaningful features. The resulting disentangled feature space provides a unified coordinate system in which target-related features specify \textbf{\textit{what to erase}} and complementary features specify \textbf{\textit{what to preserve}}. This feature-level disentanglement is particularly suitable for suppressing target concepts without disturbing the non-target generative priors.

Recent studies have explored the use of kSAEs for feature localization and suppression in concept erasure. 
However, existing approaches remain limited in achieving persistent erasure. 
Some methods conduct concept erasure using kSAEs directly during inference, without modifying any model parameters~\cite{cywinski2025saeuron,cassano2026saemnesia,he2025singleneuronworksprecise,shi2026orthoerasercoupledneuronorthogonalprojection}. The unchanged model parameters allow users to easily bypass the intervention, rendering such methods ineffective in open-source settings.
Meanwhile, some other methods operate on text representations produced by the text encoder rather than intervening in the internal visual representations of the diffusion model~\cite{tian2025sparseautoencoderzeroshotclassifier,kim2025conceptsteerersleveragingksparse}. Consequently, they merely disrupt text–concept associations without erasing the underlying visual representations, allowing the target concepts to be recovered through adversarial prompts~\cite{tsai2024ring,zhang2024defensive}.
Beyond their lack of persistence, existing kSAE-based approaches also struggle to achieve precise concept erasure.
Despite localizing target-related features within the diffusion model using kSAEs, their objectives focus solely on suppressing target features without explicitly preserving their complementary non-target ones~\cite{cywinski2025saeuron,he2025singleneuronworksprecise}, thereby failing to account for unintended changes to unrelated semantics.

To this end, we propose \textbf{PEAK}, a framework for \textbf{P}recise and persistent concept \textbf{E}r\textbf{A}sure in diffusion models via \textbf{K}-sparse autoencoders. As illustrated in Figure~\ref{fig:figure1}, PEAK addresses these limitations through kSAE-based feature localization in the internal visual space, using the disentangled features to guide parameter optimization toward precise and persistent concept erasure.
Specifically, PEAK first trains a kSAE on internal activations of the diffusion network. To precisely localize the target concept, PEAK computes sparse activations induced by prompts containing and excluding the target concept, and then contrasts both features according to their activation strength and frequency across denoising timesteps, which helps identify a compact set of target-relevant sparse features.
Subsequently, PEAK leverages the localized sparse features to guide parameter optimization, suppressing localized target activations while aligning complementary activations with their original responses in the same sparse feature space. By internalizing these erasure and preservation constraints into the model parameters, PEAK enables precise and persistent erasure under both regular and adversarial prompts without additional inference-time intervention.
Experiments show that PEAK achieves precise and persistent concept erasure, reducing the NudeNet detections from 582 to 6 on the I2P benchmark and an average attack success rate (ASR) from 96.52\% to 5.63\% on three widely used adversarial attack benchmarks, while attaining a near-zero KID on MS-COCO that indicates strong preservation of general generation quality. Our contributions:

\begin{itemize}
    \item We propose \textbf{PEAK}, which exploits the feature localization and selective suppression capabilities of kSAEs to achieve precise and persistent concept erasure.

    \item We introduce a diffusion feature localization strategy and a training objective that suppresses target-related sparse features while preserving their complementary ones.

    \item Experiments demonstrate the precision and persistence of PEAK, reducing the NudeNet detections from 582 to 6 on the I2P benchmark and an average ASR from 96.52\% to 5.63\% on three adversarial attack benchmarks, while attaining a near-zero KID on MS-COCO.
\end{itemize}
\section{Related Work}
\textbf{Concept erasure.}
Existing concept erasure methods can be broadly divided into three categories. Fine-tuning-based methods, such as ESD~\cite{gandikota2023erasing}, FMN~\cite{zhang2024forget} and CA~\cite{kumari2023ablating}, update diffusion-model parameters to suppress target concepts. Subsequent approaches further incorporate adversarial objectives or explicit preservation constraints to improve erasure persistence and reduce interference with retained concepts~\cite{srivatsan2025stereo,kim2026cooccurringassociatedretainedconcepts}. Closed-form methods directly edit selected model parameters to improve efficiency and scalability~\cite{gandikota2024unified,gong2024reliable,li2026speed}, while inference-time methods suppress undesired concepts during generation without modifying the model parameters~\cite{jain2025trascetrajectorysteeringconcept,wang2025precise}. Despite their effectiveness, these methods generally manipulate dense parameters or activations without explicitly decomposing them into target-specific and complementary features. Consequently, precisely determining \textit{what to erase} and \textit{what to preserve} remains challenging.

\noindent \textbf{SAE-based concept erasure.}
Sparse autoencoders (SAEs), including k-sparse autoencoders, provide an interpretable mechanism for decomposing dense neural activations into sparse and semantically meaningful features~\cite{huben2024sparse,bricken2023monosemanticity}. This feature-level decomposition offers a natural basis for locating and selectively suppressing concept-related representations in diffusion models. Recent studies have explored SAE-based approaches for concept localization, steering, and erasure. SAeUron~\cite{cywinski2025saeuron}, ItD~\cite{tian2025sparseautoencoderzeroshotclassifier}, and SNCE~\cite{he2025singleneuronworksprecise} leverage SAE features to identify or suppress concept-related representations during inference, while Concept Steerers~\cite{kim2025conceptsteerersleveragingksparse}, SAEmnesia~\cite{cassano2026saemnesia}, and OrthoEraser~\cite{shi2026orthoerasercoupledneuronorthogonalprojection} further investigate interpretable feature discovery and disentanglement for controllable generation or concept removal. However, existing SAE-based approaches mainly rely on inference-time feature intervention or focus on concept localization, leaving the challenge of transforming interpretable SAE features into permanent model-level concept erasure largely unexplored.
\section{Method}
\begin{figure*}[t]
\centering
\includegraphics[width=\textwidth]{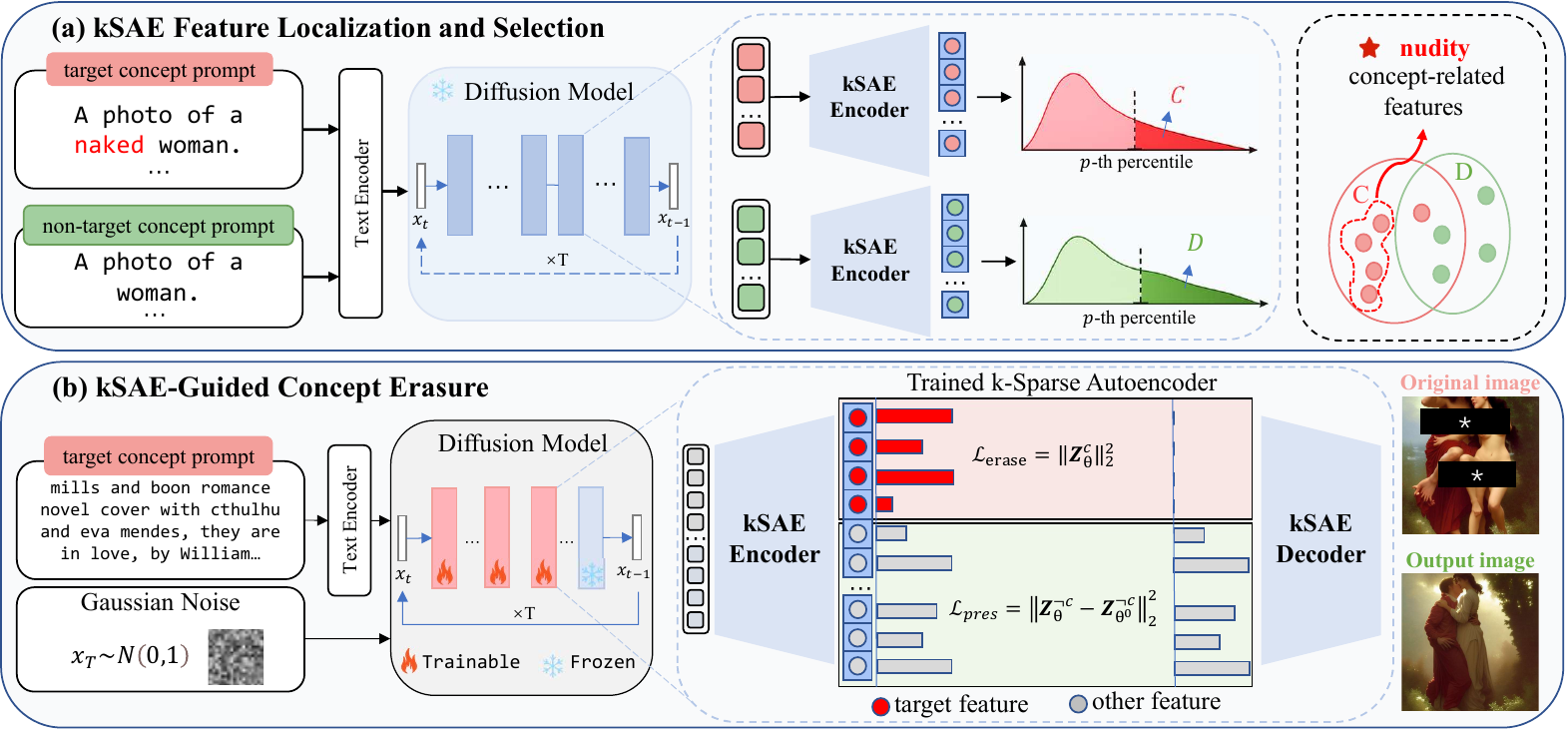}
\caption{
The main pipeline of the proposed PEAK.
(a) A frozen kSAE encodes diffusion-model activations induced by matched target and non-target prompts, and target-specific features are selected by contrasting their activation scores.
(b) The selected target features are suppressed during fine-tuning, and the other features are aligned with those of the original model.
}
\label{fig:figure2}
\end{figure*}

\textsc{PEAK} builds on a k-sparse autoencoder (kSAE) trained to decompose the internal activations of a pretrained diffusion model into sparse features. Once trained, the kSAE is frozen throughout the subsequent concept erasure process. As illustrated in Figure~\ref{fig:figure2}, concept erasure proceeds in two stages: target-specific feature localization and feature-guided parameter optimization. First, we identify the kSAE features that are strongly associated with the target concept $C$. Then, these features serve as intermediate supervision for updating the diffusion model, where target-related activations are suppressed while the complementary feature responses are aligned with those of the original model. In this way, \textsc{PEAK} embeds selective feature suppression into the model parameters while preserving non-target semantics and the model's original generative capability.

\subsection{Training kSAE for Diffusion Models}
We train a kSAE to construct a sparse feature space for the internal representations of the diffusion model. Specifically, we freeze the pretrained diffusion model and collect activations from a selected block of the denoising network throughout the denoising process. At timestep $t$, the extracted activation map is denoted by $\mathbf{F}_{t}\in\mathbb{R}^{h\times w\times d}$, where $h$ and $w$ are its spatial dimensions and $d$ is the channel dimension. Each vector at a spatial location represents one latent patch and is treated as an individual training sample. We flatten the activation maps over spatial locations and aggregate samples from all denoising timesteps. 

Let $\mathbf{x}\in\mathbb{R}^{d}$ denote an activation vector and let $n$ denote the latent dimension of the kSAE. Following the single-layer ReLU architecture~\cite{bricken2023monosemanticity}, we first compute the pre-sparse latent activation as
\begin{equation}
    \mathbf{a}
    =
    \operatorname{ReLU}\left(
    \mathbf{W}_{\mathrm{enc}}
    \left(\mathbf{x}-\mathbf{b}_{\mathrm{pre}}\right)
    +\mathbf{b}_{\mathrm{enc}}
    \right),
\end{equation}
where $\mathbf{W}_{\mathrm{enc}}\in\mathbb{R}^{n\times d}$, $\mathbf{b}_{\mathrm{pre}}\in\mathbb{R}^{d}$, and $\mathbf{b}_{\mathrm{enc}}\in\mathbb{R}^{n}$ are learnable parameters.

During training, we apply BatchTopK sparsity~\cite{bussmann2024batchtopksparseautoencoders}. Given a minibatch of $N$ activation vectors, we stack their pre-sparse activations into $\mathbf{A}\in\mathbb{R}^{N\times n}$ and retain the globally largest $Nk$ entries:
\begin{equation}
    \mathbf{Z}
    =
    \operatorname{BatchTopK}_{Nk}(\mathbf{A}).
\end{equation}
Compared with conventional sparsity-regularized SAEs, BatchTopK imposes an explicit activation budget, maintaining an average of $k$ active features per sample while allowing different spatial locations to use different numbers of features. Empirically, it allocates more active features to central regions and fewer to most peripheral regions, illustrating its adaptive allocation of sparse features across spatial locations (Appendix~\ref{app:BatchTopK}). This compact and controlled sparse support is particularly suitable for consistent concept-feature localization and selective suppression. Let $\mathbf{z}_i\in\mathbb{R}^{n}$ denote the sparse latent code associated with the $i$-th activation vector. The corresponding reconstruction is
\begin{equation}
    \hat{\mathbf{x}}_i
    =
    \mathbf{W}_{\mathrm{dec}}\mathbf{z}_i
    +\mathbf{b}_{\mathrm{pre}},
\end{equation}
where $\mathbf{W}_{\mathrm{dec}}\in\mathbb{R}^{d\times n}$ is the decoder matrix. After training, individual activation vectors are encoded using standard per-sample TopK sparsity~\cite{makhzani2014ksparseautoencoders,gao2025scaling}, i.e., $\mathbf{z}=\operatorname{TopK}_{k}\left( \mathbf{a}\right)$.

We optimize the kSAE using a normalized reconstruction loss together with the AuxK objective~\cite{gao2025scaling}. Let
$\bar{\mathbf{x}}=\frac{1}{N}\sum_{i=1}^{N}\mathbf{x}_i$
be the minibatch mean and
$\mathbf{e}_i=\hat{\mathbf{x}}_i-\mathbf{x}_i$
be the residual of the main BatchTopK reconstruction. AuxK masks out all non-dead latent features and applies BatchTopK to
the remaining pre-sparse activations with a global auxiliary budget of
$Nk_{\mathrm{aux}}$ entries, where
$k_{\mathrm{aux}}=
\min(\lfloor d/2\rfloor,|\mathcal{D}|)$,
and $\mathcal{D}$ denotes the set of latent features classified as
dead according to their firing history. Let $\tilde{\mathbf z}_i$ denote the auxiliary sparse code associated with the $i$-th activation vector. The corresponding auxiliary reconstruction is computed using the same decoder and bias as the main reconstruction: 
$\hat{\mathbf e}_i =
\mathbf W_{\mathrm{dec}}\tilde{\mathbf z}_i
+
\mathbf b_{\mathrm{pre}}.$
The auxiliary reconstruction is trained to predict the residual
$\mathbf e_i=\hat{\mathbf x}_i-\mathbf x_i$
of the main BatchTopK reconstruction. When fewer than $\lfloor d/2\rfloor$ dead features are available, we reduce the contribution of the auxiliary objective using
$s_{\mathrm{aux}}=
\min\left(
\frac{|\mathcal D|}{\lfloor d/2\rfloor},
1
\right).$
The training objective is then
\begin{equation}
\mathcal{L}_{\mathrm{kSAE}}
=
\underbrace{
\frac{\sum_{i=1}^{N}
\|\hat{\mathbf{x}}_i-\mathbf{x}_i\|_2^{2}}
{\sum_{i=1}^{N}
\|\mathbf{x}_i-\bar{\mathbf{x}}\|_2^{2}}
}_{\mathcal{L}_{\mathrm{FVU}}}
+
\alpha
s_{\mathrm{aux}}
\underbrace{
\frac{\sum_{i=1}^{N}
\|\hat{\mathbf{e}}_i-\mathbf{e}_i\|_2^{2}}
{\sum_{i=1}^{N}
\|\mathbf{x}_i-\bar{\mathbf{x}}\|_2^{2}}
}_{\mathcal{L}_{\mathrm{aux}}},
\end{equation}
where $\alpha$ controls the auxiliary contribution and $s_{\mathrm{aux}}$ adjusts its effective weight based on the number of available dead features. The FVU term encourages faithful reconstruction of diffusion activations, while the AuxK term reduces the number of dead features and improves latent utilization.

\subsection{kSAE Feature Selection for Concept Erasure}
\paragraph{Step 1: kSAE Feature Importance Scoring.}
The trained kSAE maps each diffusion activation into an $n$-dimensional sparse feature space. Our goal is to evaluate the relevance of each feature dimension $f\in\{1,\ldots,n\}$ to the target concept $C$. To reduce prompt-specific bias, relevant features should respond strongly to the concept prompts and remain consistently prominent throughout the denoising process.

We construct two matched prompt sets, $\mathcal{D}_{c}$ and $\mathcal{D}_{\neg c}$. The target set $\mathcal{D}_{c}$ contains prompts describing $C$, while the non-target set $\mathcal{D}_{\neg c}$ removes $C$ but retains the remaining prompt semantics. We compute feature importance separately for the two sets. Let $q\in\{c,\neg c\}$ denote the prompt-set type, and let $z_{i,t,f}^{q}$ be the spatially averaged activation of feature $f$ for the $i$-th prompt at denoising timestep $t$. We first average its activation across all prompts in $\mathcal{D}_{q}$:
\begin{equation}
Z_{t,f}^{q}
=
\frac{1}{|\mathcal{D}_{q}|}
\sum_{i=1}^{|\mathcal{D}_{q}|}
z_{i,t,f}^{q}.
\end{equation}

At each timestep, we collect positively activated features:
\begin{equation}
\mathcal{I}_{t}^{q}
=
\left\{
f\in\{1,\ldots,n\}
\mid
Z_{t,f}^{q}>0
\right\}.
\end{equation}
We retain the $k_{\mathrm{step}}$ features with the highest activations:
\begin{equation}
\mathcal{T}_{t}^{q}
=
\operatorname*{arg\,TopK}_{f\in\mathcal{I}_{t}^{q}}^{k_t^q}
Z_{t,f}^{q},
\qquad
k_t^q
=
\min\left(k_{\mathrm{step}},|\mathcal{I}_{t}^{q}|\right).
\end{equation}
Here, $k_{\mathrm{step}}$ denotes the number of prominent feature dimensions retained at each timestep and is distinct from the sparsity parameter $k$ of the kSAE.

The importance score of feature $f$ in set $q$ is defined as:
\begin{equation}
S_{f}^{q}
=
\underbrace{
\left(
\frac{1}{T}\sum_{t=1}^{T}Z_{t,f}^{q}
\right)
}_{\text{activation strength}}
\cdot
\underbrace{
\left(
\frac{1}{T}\sum_{t=1}^{T}
\mathbf{1}\!\left[f\in\mathcal{T}_{t}^{q}\right]
\right)
}_{\text{activation frequency}}.
\end{equation}
The first term measures the overall activation strength of feature $f$, while the second measures how frequently it appears among the most prominent features across denoising timesteps. Their product assigns high scores to features that are both strongly activated and consistently prominent. The resulting scores $S_f^{c}$ and $S_f^{\neg c}$ are subsequently contrasted to distinguish target-specific features from features shared with the non-target prompts. The effectiveness of combining activation strength and activation frequency is validated through an ablation study in Appendix~\ref{app:ablation} (c).

\paragraph{Step 2: Select Target Features}
The importance scores obtained in Step 1 measure feature prominence separately for the target and non-target prompt sets. Since the matched prompts retain the shared context while differing in the target concept $C$, a target-specific feature should receive a high score for $\mathcal{D}_{c}$ but not for $\mathcal{D}_{\neg c}$.

For each prompt set $q\in\{c,\neg c\}$, we define a percentile-based threshold
\begin{equation}
\tau_q
=
Q_p\left(\left\{S_f^q\right\}_{f=1}^{n}\right),
\end{equation}
where $Q_p(\cdot)$ denotes the $p$-th percentile over all $n$ feature scores. The target-specific feature set is then defined as
\begin{equation}
\mathcal{F}_{C}
=
\left\{
f\in\{1,\ldots,n\}
\;\middle|\;
S_f^{c}>\tau_c,
\quad
S_f^{\neg c}\leq\tau_{\neg c}
\right\}.
\label{eq:concept_feature_set}
\end{equation}
The first condition retains features with high concept importance, jointly determined by activation strength and timestep consistency, while the second excludes features that are also important for the shared semantics in the non-target prompts. The percentile $p$ controls the selection strictness, with a larger value producing a more selective feature set. The resulting $\mathcal{F}_{C}$ is used as the target-feature supervision in the subsequent concept erasure stage.

\begin{table*}[t]
\centering
\renewcommand{\arraystretch}{1.1}
\small
\setlength{\tabcolsep}{4pt}

\begin{tabular}{lccccccccc|ccc}
\toprule   
\multicolumn{1}{c}{\multirow{2}{*}{Method}} &
\multicolumn{9}{c|}{NudeNet Detection Results on I2P} & \multicolumn{3}{c}{MS-COCO} \\
\cmidrule(lr){2-10} \cmidrule(l){11-13}
 & Arm. & Bel. & But. & Fee. & Bre. (F) & Gen. (F) & Bre. (M) & Gen. (M) & Total & CS↑ & FID↓ & KID↓ \\
\midrule

SD v1.4 & 115 & 132 & 21 & 17 & 264 & 9 & 19 & 5 & 582 & 26.63 & - & - \\

\midrule

RACE~\cite{kim2024race} 
& 88 & 76 & 9 & 20 & 128 & 7 & 12 & 2 & 342 & 25.54 & 44.51 & \underline{0.0134} \\
ESD~\cite{gandikota2023erasing}
& 25 & 15 & 1 & 6 & 18 & 0 & 2 & 3 & 70 & 25.64 & 47.46 & 0.0602\\
UCE~\cite{gandikota2024unified}
& 20 & 31 & 4 & 0 & 76 & 1 & 0 & 6 & 138 & \underline{26.32} & 46.21 & 0.0377 \\
MACE~\cite{lu2024mace}
& 30 & 16 & 2 & 12 & 30 & 2 & 2 & 3 & 97 & 24.04 & 54.09 & 0.1927 \\
RECE~\cite{gong2024reliable}
& 12 & 16 & 2 & 4 & 13 & 1 & 5 & 1 & 54 & 26.14 & \underline{42.19} & \textbf{0.0000} \\
AdvUn~\cite{zhang2024defensive}
& 7 & 6 & 0 & 2 & 4 & 0 & 1 & 1 & 21 & 23.96 & 49.11 & 0.0563 \\
TraSCE~\cite{jain2025trascetrajectorysteeringconcept}
& 5 & 2 & 1 & 3 & 4 & 0 & 0 & 0 & 15 & 25.05 & 69.53 & 0.5443 \\
ReCARE~\cite{kim2026cooccurringassociatedretainedconcepts}
& 2 & 3 & 1 & 0 & 0 & 0 & 0 & 1 & \underline{7} & 25.53 & 50.56 & 0.2096 \\
STEREO~\cite{srivatsan2025stereo} & 2 & 2 & 1 & 0 & 0 & 0 & 1 & 0 & \textbf{6} & 25.16 & 54.24 & 0.2919 \\

\midrule
\textbf{PEAK} (Ours) & 0 & 0 & 2 & 2 & 1 & 0 & 0 & 1 & \textbf{6} & \textbf{26.46} & \textbf{41.59} & \textbf{0.0000} \\
\bottomrule   
\end{tabular}
\caption{
Evaluation of implicit concept erasure on the I2P benchmark.
We report NudeNet detections at a confidence threshold of 0.6.
Here, Arm., Bel., But., Fee., Bre., and Gen. denote Armpits, Belly, Buttocks, Feet, Breasts, and Genitalia, respectively, with (F) and (M) indicating Female and Male categories.
Best and second-best results are bolded and underlined.
}
\label{tab:nudenet}
\end{table*}
\subsection{kSAE-Guided Concept Erasure}
After obtaining the selected target-specific kSAE feature set $\mathcal{F}_C$, we use these features as intermediate supervision to fine-tune the diffusion model. Specifically, the selected target features are suppressed, while their complementary feature responses are aligned with those of the original model. These two constraints jointly determine \textit{what to erase} and \textit{what to preserve} in the same sparse feature space.

\paragraph{On-Trajectory Sparse Feature Extraction.}
Let $\Theta_{0}$ denote the frozen parameters of the original denoising network, and let $\Theta$ denote a trainable copy initialized from $\Theta_{0}$. The frozen kSAE encoder is denoted by $E$. For each target-concept training prompt $\mathbf{y}$, we sample Gaussian noise $\mathbf{x}_{T}\sim\mathcal{N}(\mathbf{0},\mathbf{I})$ and randomly sample a rollout depth $r$. The frozen reference model performs $r$ denoising steps to obtain
\begin{equation}
\widetilde{\mathbf{x}}_{t_r}
=
\mathcal{R}^{r}_{\Theta_{0}}
\left(\mathbf{x}_{T},\mathbf{y}\right),
\end{equation}
where $\mathcal{R}^{r}_{\Theta_{0}}$ denotes the corresponding partial denoising trajectory and $t_r$ is the resulting timestep. This rollout produces a latent state that is actually visited by the original model under the target-concept condition, providing more representative supervision than a noisy latent sampled independently of the generation trajectory.

Given the same prompt $\mathbf{y}$, timestep $t_r$, and latent state $\widetilde{\mathbf{x}}_{t_r}$, we extract the activations of the trainable and reference models at the selected denoising-network layer $\ell$ and encode them into the kSAE feature space:
\begin{equation}
\mathbf{Z}_{\psi}
=
E\left(
h_{\psi}^{\ell}
\left(
\widetilde{\mathbf{x}}_{t_r},
t_r,
\mathbf{y}
\right)
\right),
\qquad
\psi\in\{\Theta,\Theta_{0}\}.
\end{equation}
Here, $h_{\psi}^{\ell}(\cdot)$ denotes the activation at layer $\ell$ under parameters $\psi$. The resulting sparse feature tensors
$\mathbf{Z}_{\Theta},\mathbf{Z}_{\Theta_{0}}
\in\mathbb{R}^{B\times M\times n}$
correspond to the trainable and reference models, respectively, where $B$ is the batch size, $M$ is the number of spatial locations, and $n$ is the kSAE feature dimension.

\paragraph{Target-Feature Suppression.}
To erase the target concept, we minimize the activation energy of the selected target-related features $\mathcal{F}_{C}$:
\begin{equation}
\mathcal{L}_{\mathrm{erase}}(\Theta)
=
\frac{1}{BM|\mathcal{F}_{C}|}
\sum_{b=1}^{B}
\sum_{m=1}^{M}
\sum_{f\in\mathcal{F}_{C}}
\left(
[\mathbf{Z}_{\Theta}]_{b,m,f}
\right)^2.
\end{equation}
Minimizing $\mathcal{L}_{\mathrm{erase}}$ drives the target-specific sparse activations toward zero, weakening the internal features responsible for generating the target concept.

\paragraph{Complementary-Feature Preservation.}
We find that suppressing the target features alone may inadvertently alter other internal representations. To mitigate this issue, we preserve the complementary feature set 
\begin{equation}
\overline{\mathcal{F}}_{C}
=
\{1,\ldots,n\}\setminus\mathcal{F}_{C},
\end{equation}
by matching responses to the frozen reference model:
\begin{equation}
\small
\mathcal{L}_{\mathrm{pres}}(\Theta)
=
\frac{1}{BM|\overline{\mathcal{F}}_{C}|}
\sum_{b=1}^{B}
\sum_{m=1}^{M}
\sum_{f\in\overline{\mathcal{F}}_{C}}
\left(
[\mathbf{Z}_{\Theta}]_{b,m,f}
-
[\mathbf{Z}_{\Theta_{0}}]_{b,m,f}
\right)^2.
\end{equation}
This feature-space distillation constrains the trainable model to retain the complementary sparse representations of the original model, reducing interference with non-target semantics during fine-tuning.

\paragraph{Overall Objective.}
The final objective is
\begin{equation}
\mathcal{L}_{\mathrm{PEAK}}(\Theta)
=
\mathcal{L}_{\mathrm{erase}}(\Theta)
+
\lambda\mathcal{L}_{\mathrm{pres}}(\Theta),
\end{equation}
where the hyperparameter $\lambda$ balances target-feature suppression and complementary-feature preservation. During optimization, $\Theta_{0}$ and $E$ remain frozen, and only selected parameters in the trainable denoising network $\Theta$ is updated. After fine-tuning, neither the reference model nor the kSAE is required during inference, since the feature-level erasure has been embedded into the diffusion-model parameters.
\section{Experiments}
\begin{table*}[ht]
\centering
\renewcommand{\arraystretch}{1.1}
\small
\setlength{\tabcolsep}{6pt}

\begin{tabular}{lcccc|c|cc|ccc}
\toprule
\multicolumn{1}{c}{\multirow{2}{*}{Method}}
& \multicolumn{4}{c|}{RAB $\downarrow$} 
& \multirow{2}{*}{MMA $\downarrow$} 
& \multicolumn{2}{c|}{UnlearnDiffAtk $\downarrow$} 
& \multicolumn{3}{c}{MS-COCO} \\
\cmidrule(lr){2-5} \cmidrule(lr){7-8} \cmidrule(l){9-11}
& K16 & K38 & K77 & AVG 
& 
& Pre-ASR & Post-ASR 
& CS$\uparrow$ & FID$\downarrow$ & KID$\downarrow$ \\
\midrule

SD1.4 
& 93.68 & 97.89 & 92.63 & 94.74 
& 96.50 
& 88.14 & 98.31 
& 26.63 & - & - \\

\midrule

AdvUn~\cite{zhang2024defensive} 
& \underline{1.05} & \textbf{0.00} & \underline{1.05} & \underline{0.70} 
& \textbf{0.30} 
& 6.78 & 16.95 
& 23.96 & 49.11 & 0.0563 \\

RACE~\cite{kim2024race}
& 90.53 & 93.68 & 98.95 & 94.39 
& 43.90 
& 62.71 & 96.61 
& 25.54 & 44.51 & \underline{0.0134} \\

RECE~\cite{gong2024reliable}
& 14.74 & 24.21 & 14.74 & 17.89 
& 23.00 
& 13.56 & 58.47 
& 26.14 & \underline{42.19} & \textbf{0.0000} \\

UCE~\cite{gandikota2024unified}
& 17.89 & 18.95 & 11.58 & 16.14 
& 24.20 
& 25.42 & 86.44 
& \underline{26.32} & 46.21 & 0.0377 \\

ESD~\cite{gandikota2023erasing}
& 34.74 & 42.11 & 45.26 & 40.70 
& 6.70 
& 16.95 & 83.05 
& 25.64 & 47.46 & 0.0602 \\

MACE~\cite{lu2024mace}
& \underline{1.05} & \underline{2.11} & 2.11 & 1.75 
& 1.40 
& 8.47 &  73.73
& 24.04 & 54.09 & 0.1927 \\

ReCARE~\cite{kim2026cooccurringassociatedretainedconcepts}
& \underline{1.05} & 3.16 & 3.16 & 2.46 
& 3.00 
& \underline{4.24} & 41.53 
& 25.53 & 50.56 & 0.2096 \\

STEREO~\cite{srivatsan2025stereo}
& \underline{1.05} & 3.16 & \underline{1.05} & 1.40 
& 1.10 
& \underline{4.24} & \textbf{15.25} 
& 25.16 & 54.24 & 0.2919 \\

\midrule

\textbf{PEAK} (Ours) 
& \textbf{0.00} & \textbf{0.00} & \textbf{0.00} & \textbf{0.00} 
& \underline{0.80} 
& \textbf{0.85} & \underline{16.10} 
& \textbf{26.46} & \textbf{41.59} & \textbf{0.0000} \\

\bottomrule
\end{tabular}

\caption{
Robustness and preservation comparison under adversarial concept-recovery attacks. RAB, MMA, and UnlearnDiffAtk report attack success rates, while CS, FID, and KID evaluate generation preservation on MS-COCO. All ASRs are reported in percentages. Lower ASR, FID, and KID values and higher CS values are better.
}
\label{tab:adv_attack}
\end{table*}
\subsection{kSAE Experimental Setup}
\paragraph{Where to apply kSAE.} Previous mechanistic studies~\citep{basu2024localizing,basu2024mechanistic} show that cross-attention blocks in different U-Net stages specialize in controlling distinct visual attributes. Based on this observation, we apply kSAE to cross-attention outputs in the U-Net upsampling path and perform block-wise ablation. We select \texttt{up.1.2} for style erasure and \texttt{up.1.1} for object and nudity erasure, which provide the most effective intervention locations. Qualitative comparisons are provided in Appendix~\ref{app:where_sae}.

\paragraph{kSAE training data and activation collection.} 
The kSAE is trained on diverse prompts to learn sparse representations of diffusion model activations. The training set includes general image-caption prompts from MS-COCO~\cite{lin2014microsoft} and inappropriate-content prompts from a separate split of I2P~\cite{schramowski2023safe}. We further include template-generated prompts covering representative object and artistic-style concepts to improve the diversity of the learned sparse features. Detailed training configurations are provided in Appendix~\ref{app:SAE trainings details}. 

\subsection{Interpreting kSAE Features}
We examine whether the selected kSAE features are both discriminative of and semantically related to the target concept.

\begin{figure}[t]
\centering
\includegraphics[width=\linewidth]{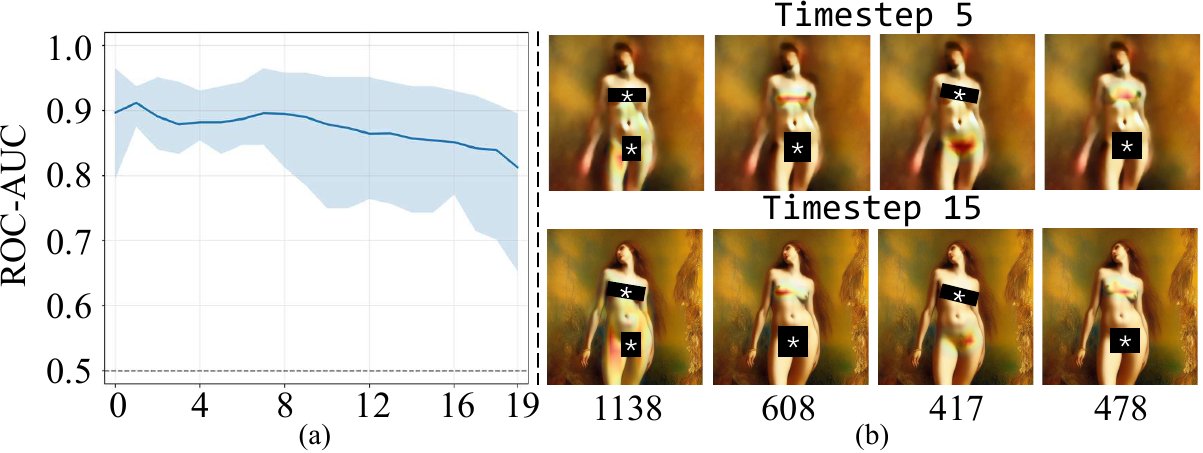}
\caption{
(a) The selected features maintain high ROC-AUC across denoising timesteps.
(b) Activation maps of selected kSAE features, where the labels denote feature indices.
}
\label{fig:sae_analysis}
\end{figure}

\paragraph{Q1: Are the selected kSAE features discriminative?}
We evaluate the discriminative ability of selected kSAE features using directional ROC-AUC~\cite{hanley1982meaning} between nudity and matched non-target prompts. As shown in Figure~\ref{fig:sae_analysis} (a), the mean AUC remains above $0.8$ across denoising timesteps, indicating strong and temporally consistent separation between target and non-target activations. Detailed results are provided in Appendix~\ref{app:feature_discriminative_analysis}.

\paragraph{Q2: Are the discriminative features related to the target concept?}
Discriminability alone may arise from incidental contextual cues. We therefore reshape each feature's spatial activations into a heatmap and overlay it on the generated image. As shown in Figure~\ref{fig:sae_analysis} (b), the selected features consistently focus on target-related body regions. This spatial alignment suggests that the selected kSAE features capture localized and semantically meaningful evidence of the target concept. Their functional relevance is further examined through the subsequent feature-intervention study.

\subsection{Main Results}

We evaluate \textsc{PEAK} on nudity, object, and artistic-style concepts, jointly assessing target erasure and non-target preservation. For nudity, we additionally examine persistence against black- and white-box concept-recovery attacks.

\subsubsection{(1) Nudity Concept Erasure}
\paragraph{Comparison and analysis.}
We first evaluate nudity erasure on Stable Diffusion v1.4~\cite{StableDiffusion} using the I2P~\cite{schramowski2023safe} benchmark. I2P contains 4,703 implicit inappropriate prompts covering nudity content and violence. We focus on nudity erasure in the main evaluation and additionally provide a qualitative analysis of violence-related features in Appendix~\ref{app:violence}. For each prompt, we generate one image and use NudeNet~\cite{bedapudi2019nudenet} with a confidence threshold of $0.6$ to detect exposed body parts. We report category-wise and total detection counts, where a lower value indicates more effective nudity erasure. To evaluate non-target preservation, we additionally generate images from 1,000 prompts sampled from MS-COCO. We report CLIP Score (CS)~\cite{radford2021learning}, Fr\'echet Inception Distance (FID)~\cite{heusel2017gans}, and Kernel Inception Distance (KID)~\cite{inproceedings}. A higher CS indicates better prompt-image alignment, while lower FID and KID indicate better distributional quality. For readability, all KID values are multiplied by $100$. Additional evaluations on SDXL~\cite{ICLR2024_081b0806} and FLUX~\cite{flux2024} further validate the generality of \textsc{PEAK} (Appendix~\ref{app:nudity_sdxl_flux}).

We compare \textsc{PEAK} against nine concept erasure methods, including RACE, ESD, UCE, MACE, RECE, AdvUn, ReCARE, TraSCE and STEREO. As shown in Table~\ref{tab:nudenet}, \textsc{PEAK} reduces the NudeNet detection count from \textbf{582} to \textbf{6}. More importantly, \textsc{PEAK} achieves the strongest overall preservation performance, with the highest CS of \textbf{26.46}, the lowest FID of \textbf{41.59}, and a KID of \textbf{0}, which ties the best result. Compared with existing methods, \textsc{PEAK} achieves a better balance between target erasure and generation quality.

\begin{figure}[t]
\centering
\includegraphics[width=\columnwidth]{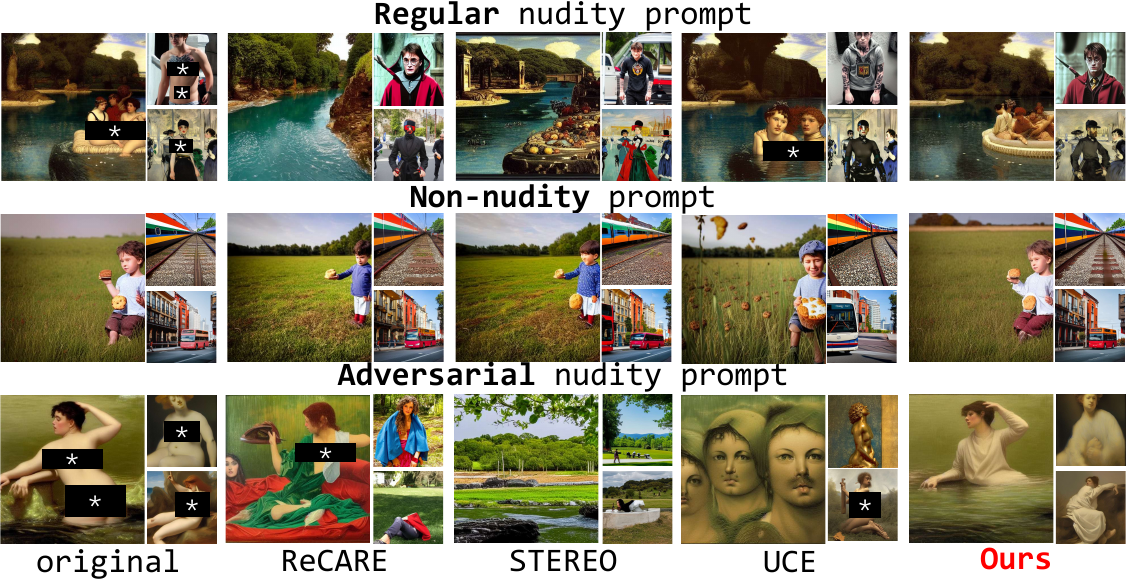}
\caption{
Qualitative comparison of different prompt types.
PEAK removes nudity while preserving non-nudity quality.}
\label{fig:contrast}
\end{figure}

\paragraph{Persistence under concept-recovery attacks.}
We further evaluate whether the erased nudity concept can be recovered through adversarial prompts. We select RAB~\cite{tsai2024ring} and MMA~\cite{yang2024mma} as black-box attacks and UnlearnDiffAtk~\cite{zhang2024defensive} as a white-box attack. As reported in Table~\ref{tab:adv_attack}, \textsc{PEAK} achieves zero successful recoveries under RAB and an MMA ASR of 0.8\%. Under UnlearnDiffAtk, it obtains the best Pre-ASR (\textbf{0.85\%}) and the second-best Post-ASR (\textbf{16.10\%}). Together with its leading COCO preservation results, \textsc{PEAK} maintains persistent erasure without substantial degradation of non-target generation. Figure~\ref{fig:contrast} provides consistent qualitative evidence under regular, non-target and adversarial prompts.

\subsubsection{(2) Object and Style Concept Erasure}
\paragraph{Comparison and analysis.}
Following~\cite{lyu2024one}, we evaluate erasure of ``Snoopy'' and ``Van Gogh'' using $80$ object and $30$ style prompt templates, respectively, with $10$ images per template. A CLIP-based zero-shot classifier~\cite{radford2021learning} measures target leakage ($\mathrm{Acc}_{e}\!\downarrow$) and non-target preservation ($\mathrm{Acc}_{u}\!\uparrow$). As shown in Table~\ref{tab:object_style}, \textsc{PEAK} achieves $\mathrm{Acc}_{e}=0$ and $\mathrm{Acc}_{u}=94.84$ for Snoopy. For Van Gogh, it obtains $\mathrm{Acc}_{e}=13.67$ and the highest $\mathrm{Acc}_{u}=81.58$. Although TraSCE yields lower Van Gogh leakage ($1.00$), its preservation accuracy drops to $57.17$, indicating substantially greater collateral degradation. Overall, \textsc{PEAK} provides the most consistent erasure-preservation trade-off across concept types. We further report multi-concept erasure and unseen-concept generalization to ``Mario'' in Appendices~\ref{app:multi_object} and~\ref{app:ksae_transferable}, respectively.
\begin{table}[t]
\centering
\label{tab:object_style_erasure}
\begin{tabular}{lcccc}
\toprule
\multirow{2}{*}{Method}
& \multicolumn{2}{c}{Erase ``Snoopy''}
& \multicolumn{2}{c}{Erase ``Van Gogh''} \\
\cmidrule(lr){2-3}
\cmidrule(lr){4-5}
& $\mathrm{Acc}_e \downarrow$
& $\mathrm{Acc}_u \uparrow$
& $\mathrm{Acc}_e \downarrow$
& $\mathrm{Acc}_u \uparrow$ \\
\midrule
Original & 94.75 & 94.16 & 95.00 & 79.83 \\
\midrule
CA      & 5.88 & 87.59 & 56.00 & 74.92 \\
MACE    & \underline{0.38} & 53.06 & 17.33 & \underline{79.08} \\
UCE  & \underline{0.38} & \textbf{95.13} & 14.00 & 73.25 \\
TraSCE  & \underline{0.38} & 68.09 & \textbf{1.00} & 57.17 \\
\textbf{PEAK} (Ours)    & \textbf{0.00} & \underline{94.84} & \underline{13.67}    & \textbf{81.58}    \\
\bottomrule
\end{tabular}
\caption{Object and artistic-style concept erasure results for ``Snoopy'' and ``Van Gogh''. Lower $\mathrm{Acc}_e$ and higher $\mathrm{Acc}_u$ indicate better erasure and preservation, respectively.}
\label{tab:object_style}
\end{table}
\subsection{Ablation Study}

\paragraph{Effect of kSAE Feature Localization.}
We further examine whether localized kSAE features control specific nudity attributes. Based on their activation patterns, features 1138/861, 864, and 608/478/756 are associated with armpit, belly, and breast attributes, respectively. Zeroing each feature group reduces the corresponding NudeNet detections by 94, 95, and 276, confirming attribute-specific control. Jointly suppressing all selected features further reduces detections to 4, 1, and 2, showing that these features provide complementary control over the target concept. Additional ablations are provided in Appendix~\ref{app:ablation}.
\begin{table}[t]
\centering
\small
\setlength{\tabcolsep}{2.2pt}
\renewcommand{\arraystretch}{1.15}
\begin{tabular*}{\columnwidth}{
    @{\extracolsep{\fill}}
    c c c c c c c c c
    @{}
}
\toprule
\multicolumn{6}{c}{Feature}
& \multicolumn{3}{c}{NudeNet Detections $\downarrow$} \\
\cmidrule(lr){1-6}
\cmidrule(lr){7-9}
1138 & 861 & 864 & 608 & 478 & 756
& Armpits & Belly & Breasts \\
\midrule
$\times$ & $\times$ & $\times$ & $\times$ & $\times$ & $\times$
& 115 & 132 & 283 \\
\midrule
$\checkmark$ & $\checkmark$ & $\times$ & $\times$ & $\times$ & $\times$
& $\textbf{21}\,(\textbf{-94})$ & 90 & 194 \\

$\times$ & $\times$ & $\checkmark$ & $\times$ & $\times$ & $\times$
& 102 & $\textbf{37}\,(\textbf{-95})$ & 250 \\

$\times$ & $\times$ & $\times$ & $\checkmark$ & $\checkmark$ & $\checkmark$
& 38 & 52 & $\textbf{7}\,(\textbf{-276})$ \\
\midrule
$\checkmark$ & $\checkmark$ & $\checkmark$
& $\checkmark$ & $\checkmark$ & $\checkmark$
& $\textbf{4}\,(\textbf{-111})$
& $\textbf{1}\,(\textbf{-131})$
& $\textbf{2}\,(\textbf{-281})$ \\
\bottomrule
\end{tabular*}
\caption{
Feature intervention on localized nudity-related kSAE features. Checkmarks denote zeroed features, bold values the most affected NudeNet category, and parentheses reductions from baseline.
}
\label{tab:ablation}
\end{table}
\section{Conclusion}
In this work, we present \textsc{PEAK}, a feature-guided approach for precise and persistent concept erasure in diffusion models. By identifying concept-specific features within internal diffusion representations, \textsc{PEAK} selectively removes target concepts while preserving unrelated semantics and overall generation quality. \textsc{PEAK} uses the localized sparse features to explicitly guide model parameter optimization, suppressing target features while constraining changes to non-target features to reduce unintended effects on unrelated semantics. Extensive experiments demonstrate effective erasure across unsafe, object, and artistic-style concepts, as well as strong robustness against adversarial prompts and consistent generalization across different diffusion architectures. These results highlight the potential of interpretable feature-level manipulation for building safer, more reliable, and controllable generative models.
\clearpage

\bibliography{reference}

\clearpage
\appendix

\clearpage

\section{Ablation Studies}
\label{app:ablation}
Throughout Tables~\ref{tab:ablation_lambda}--\ref{tab:feature_selection}, RAB and MMA report the numbers of successful concept recoveries over 285 and 1,000 attack trials, respectively, rather than attack success rates. Lower values indicate stronger erasure robustness.
\subsubsection{(a) Effect of Preservation Weight.}
We investigate the effect of the preservation weight $\lambda$, which controls the strength of non-target feature preservation during model optimization. As shown in Table~\ref{tab:ablation_lambda}, removing the preservation constraint ($\lambda=0$) results in severe degradation of generation quality. Increasing $\lambda$ improves prior preservation by constraining unnecessary changes to non-target representations. However, an overly large preservation weight may weaken concept erasure due to excessive restrictions on model updates. We therefore use $\lambda=0.02$, which provides the best erasure–preservation trade-off.
\begin{table}[ht]
\centering
\small
\setlength{\tabcolsep}{3.5pt}
\begin{tabular}{cccccc}
\toprule
$\lambda$ & I2P $\downarrow$ & RAB $\downarrow$ & MMA $\downarrow$ & CS $\uparrow$ & FID $\downarrow$ \\
\midrule
Original 
& 582 & 270 & 965 & 26.63 & - \\
\midrule
0.00 & 0  & 0 & 0  & 16.64 & 184.85 \\
0.01 & 9  & 0 & 5  & 26.26 & 44.10 \\
0.03 & 10 & 1 & 7  & 26.37 & 40.27 \\
0.04 & 10 & 0 & 8  & 26.43 & 39.12 \\
0.05 & 9  & 2 & 12 & 26.44 & 38.13 \\
\midrule
0.02 & 6 & 0 & 8 & 26.46 & 41.59 \\
\bottomrule
\end{tabular}
\caption{Ablation study on the preservation weight $\lambda$.}
\label{tab:ablation_lambda}
\end{table}

\subsubsection{(b) Effect of Parameter Updating Strategy.}
We compare four parameter updating strategies:
\textit{esd-x-strict}, which updates only the key and value projections in
cross-attention layers; \textit{esd-x}, which updates all cross-attention
parameters; \textit{esd-u}, which updates non-cross-attention parameters; and
\textit{esd-all}, which updates the entire U-Net. As shown in
Table~\ref{tab:ablation_para}, \textit{esd-x} provides the best balance
between concept erasure and generation quality. Therefore, we use \textit{esd-x} as the default strategy in PEAK.
\begin{table}[ht]
\centering
\small
\setlength{\tabcolsep}{3.5pt}
\begin{tabular}{cccccc}
\toprule
Para. & I2P $\downarrow$ & RAB $\downarrow$ & MMA $\downarrow$ & CS $\uparrow$ & FID $\downarrow$ \\
\midrule
Original 
& 582 & 270 & 965 & 26.63 & - \\
\midrule
esd-x-strict 
& 12 & 1 & 4 & 25.68 & 47.23 \\
esd-u 
& 18 & 7 & 24 & 26.50 & 34.97 \\
esd-all 
& 14 & 4 & 21 & 26.56 & 35.94 \\
\midrule
esd-x 
& 6 & 0 & 8 & 26.46 & 41.59 \\
\bottomrule
\end{tabular}
\caption{Ablation study on the parameter updating strategy.}
\label{tab:ablation_para}
\end{table}

\subsubsection{(c) Effect of Feature Selection Strategies.}
We ablate activation strength and timestep consistency by comparing four
feature-selection strategies: random, strength only, frequency only, and their combination. All variants select seven SAE features under identical training and evaluation settings. As shown in Table~\ref{tab:feature_selection}, random features fail
to erase nudity. Both individual criteria substantially improve erasure, but strength-only is less robust to MMA, while frequency-only degrades
generation quality (FID 47.98). Their combination achieves the best erasure performance, with only 6 I2P detections and zero RAB detections, while retaining competitive generation quality.
\begin{table}[ht]
\centering
\small
\setlength{\tabcolsep}{3pt}
\renewcommand{\arraystretch}{1.3}
\begin{tabular}{cc|ccccc}
\toprule
\multicolumn{2}{c|}{Feature Selection}
& \multirow{2}{*}{I2P $\downarrow$} 
& \multirow{2}{*}{RAB $\downarrow$}
& \multirow{2}{*}{MMA $\downarrow$}
& \multirow{2}{*}{CS $\uparrow$}
& \multirow{2}{*}{FID $\downarrow$}
\\
\cmidrule(lr){1-2}
Strength
& Frequency
& & & & & 
\\
\midrule
\multicolumn{2}{c|}{Original} 
& 582 & 270 & 965 & 26.63 & -
\\
\midrule
\multicolumn{2}{c|}{Random}
& 619 & 231 & 470 & 26.48 & 40.93
\\

$\checkmark$ & $\times$
& 8 & 1 & 16 & 26.53 & 41.64
\\

$\times$ & $\checkmark$
& 9 & 5 & 4 & 26.18 & 47.98
\\
\midrule
$\checkmark$ & $\checkmark$
& 6 & 0 & 8 & 26.46 & 41.59
\\
\bottomrule
\end{tabular}
\caption{Ablation study on the feature selection strategy.}
\label{tab:feature_selection}
\end{table}

\section{Additional Analysis of kSAE Representations}
\subsection{BatchTopK kSAE trained for diffusion models}
\label{app:BatchTopK}
\begin{figure}[t]
    \centering
    \includegraphics[width=\linewidth]{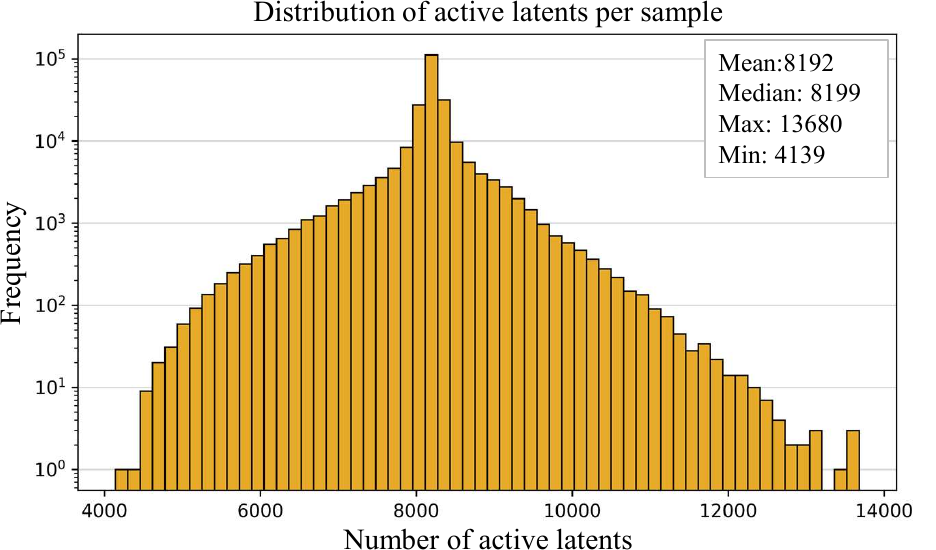}
    \caption{Number of active features per image sample. The kSAE assigns different numbers of active features to different samples, indicating that samples require different sparse budgets to achieve low reconstruction error.
    }
    \label{fig:Distribution of active latents per sample}
\end{figure}

The BatchTopK variant of kSAE enables the model to flexibly distribute active features across a data batch to achieve better reconstruction performance. Specifically, our kSAE allocates more active latents to image patches with detailed content, while less important areas, such as the background, are reconstructed using fewer features. As shown in Figure~\ref{fig:Distribution of active latents per sample}, the kSAE distributes active features unevenly across image samples. While most of the distribution centers around a mean of 8192 (since k = 32 and each image contains 16 × 16 activation vectors), a notable number of samples use significantly fewer or more active features.

\begin{figure}[t]
    \centering
    \includegraphics[width=\linewidth]{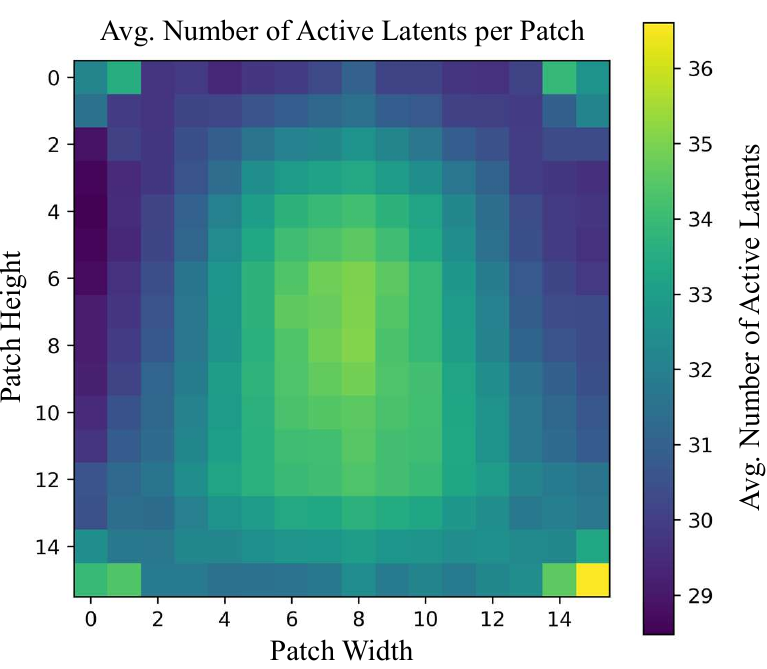}
    \caption{Average number of active features for different image patches. The BatchTopK kSAE allocates more active features to activation vectors corresponding to central image regions.}
    \label{fig:Number of Active Latents per Patch}
\end{figure}

Additionally, Figure~\ref{fig:Number of Active Latents per Patch} shows the average number of activated features per image patch. Central regions of the image tend to have more active features, while background areas have fewer. Interestingly, corners of the images also exhibit frequent activations.

\subsection{Selection of internal blocks for kSAE training}
\label{app:where_sae}
\begin{figure}[ht]
    \centering
    \includegraphics[width=\linewidth]{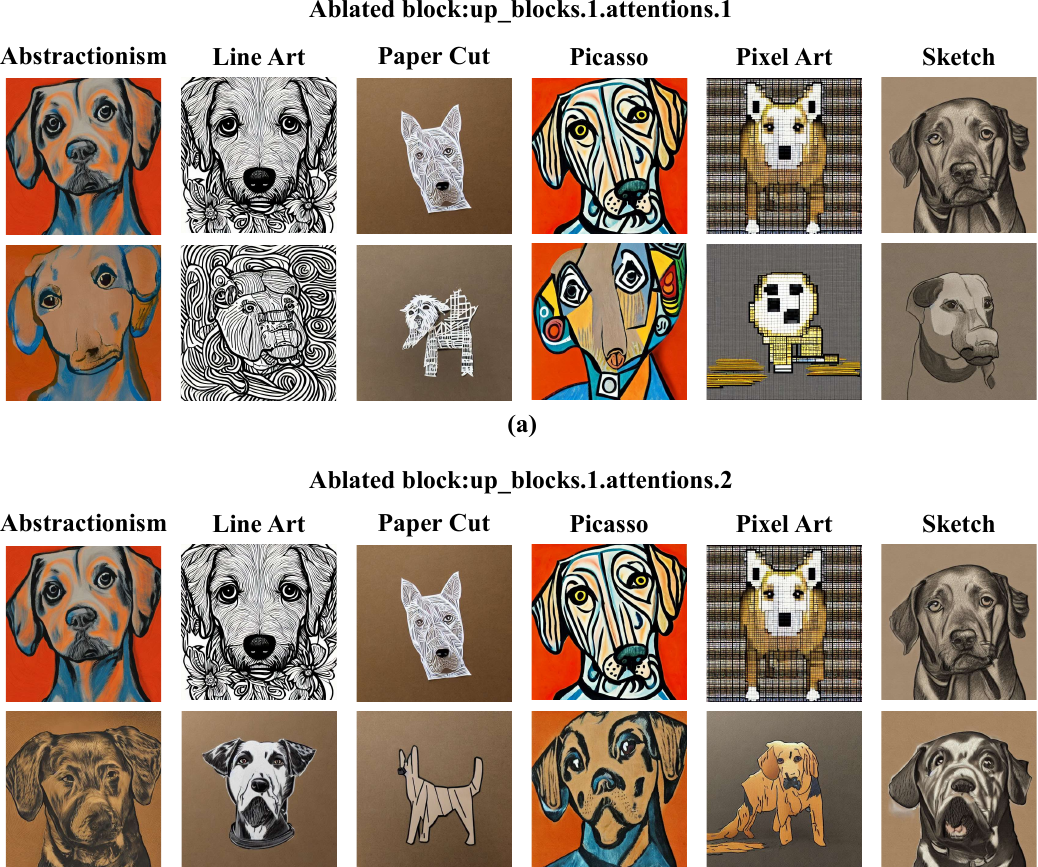}
    \caption{(a) Ablating the object block up.1.1 notably degrades the quality of generated objects and (b) ablating the style block up.1.2 almost completely removes the original style of the image.}
    \label{fig:where_sae}
\end{figure}

In LLMs, kSAEs are typically trained on activations from the residual stream, MLP layers, or attention layers~\cite{kissane2024interpretingattentionlayeroutputs}. Following recent mechanistic studies on diffusion models~\cite{basu2024localizing}, we apply kSAE to cross-attention blocks. We identify suitable blocks through block-wise ablation, where each cross-attention block is replaced with an identity function and the block causing the largest degradation of the target attribute is selected. As shown in Figure~\ref{fig:where_sae} (a) and (b), up.1.1 and up.1.2 are selected for object and style concepts, respectively. We further perform the same analysis on FLUX.1-dev, a DiT-based diffusion model~\cite{flux2024}, and select the most influential transformer block for kSAE training.
\begin{figure}[ht]
    \centering
    \includegraphics[width=\linewidth]{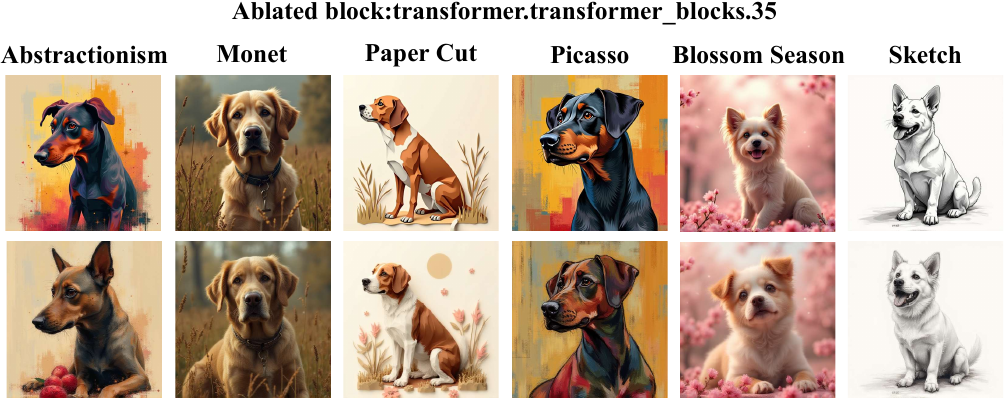}
    \caption{
    Ablating transformer block 35 affects object generation, motivating its selection for kSAE feature localization.
    }
    \label{fig:where_sae_flux}
\end{figure}

\subsection{Feature-wise Discriminative Analysis of Selected kSAE Features}
\label{app:feature_discriminative_analysis}
To further examine whether the selected kSAE features are truly associated with the target concept, we evaluate their feature-wise discriminative ability between nudity prompts and no-nudity prompts. Specifically, for each selected feature and each denoising timestep, we use the corresponding kSAE activation as a scalar classification score and compute a directional ROC-AUC. Nudity prompts are treated as the positive class, while no-nudity prompts are treated as the negative class.

Let $z^{c}_{i,t,f}$ denote the activation score of feature $f$ at denoising timestep $t$ for the $i$-th nudity prompt, and let $z^{\neg c}_{j,t,f}$ denote the corresponding activation score for the $j$-th no-nudity prompt. Here, $N_c$ and $N_\neg c$ denote the numbers of nudity and no-nudity prompts, respectively. The directional ROC-AUC is computed as
\begin{equation}
\small
\operatorname{AUC}_{t,f}
=
\frac{1}{N_c N_\neg c}
\sum_{i=1}^{N_c}
\sum_{j=1}^{N_\neg c}
\left[
\mathbb{I}\left(z^{c}_{i,t,f} > z^{\neg c}_{j,t,f}\right)
+
\frac{1}{2}
\mathbb{I}\left(z^{c}_{i,t,f} = z^{\neg c}_{j,t,f}\right)
\right],
\label{eq:directional_auc}
\end{equation}
where $\mathbb{I}(\cdot)$ is the indicator function. This rank-based formulation is equivalent to the standard ROC-AUC interpretation, the probability that a randomly sampled positive example receives a higher score than a randomly sampled negative example~\cite{hanley1982meaning}. Therefore, $\operatorname{AUC}_{t,f}=0.5$ indicates no discriminative ability, while $\operatorname{AUC}_{t,f}>0.5$ indicates that feature $f$ tends to activate more strongly for nudity prompts than for non-nudity prompts at timestep $t$.

\begin{figure}[ht]
    \centering
    \includegraphics[width=\linewidth]{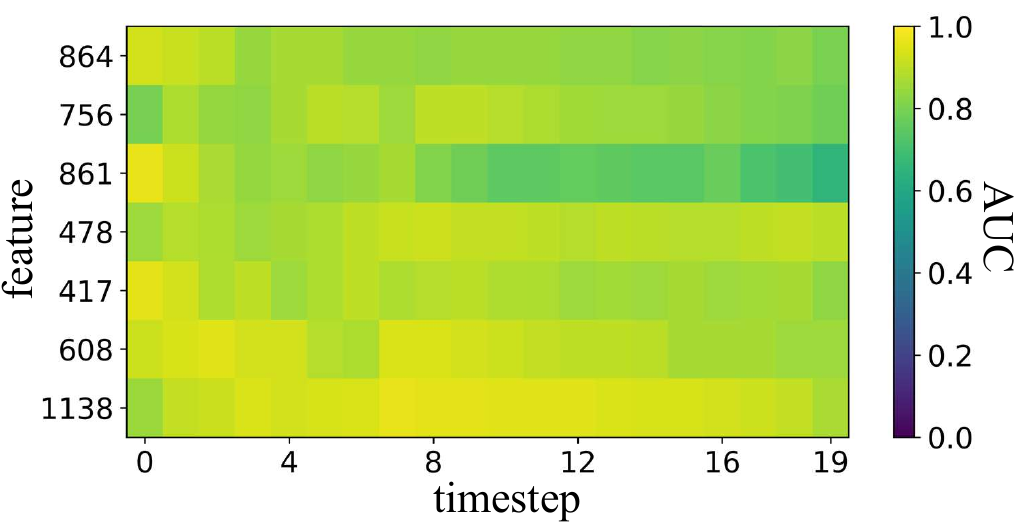}
    \caption{
        Feature-wise ROC-AUC analysis of selected kSAE features.
        Columns correspond to selected features and rows to denoising timesteps. 
        Higher AUC values indicate stronger discrimination between nudity and non-nudity prompts.
    }
    \label{fig:features_auc_heatmap}
\end{figure}

Figure~\ref{fig:features_auc_heatmap} visualizes the resulting feature-wise AUC matrix. Each row corresponds to one selected nudity-related kSAE feature, and each column corresponds to a denoising timestep. The heatmap shows that most selected features consistently maintain high directional AUC values across the generation trajectory. This demonstrates that the selected features are not merely active at isolated timesteps, but encode stable target concept discriminative signals throughout the denoising process.

\section{Additional Evaluations}
\subsection{Cross-Architecture Concept Erasure}
\label{app:nudity_sdxl_flux}
To further demonstrate the generality of PEAK beyond the default Stable Diffusion setting, we evaluate its concept erasure capability on different diffusion architectures, including SDXL and FLUX.

\begin{figure}[!ht]
    \centering
    \includegraphics[width=\linewidth]{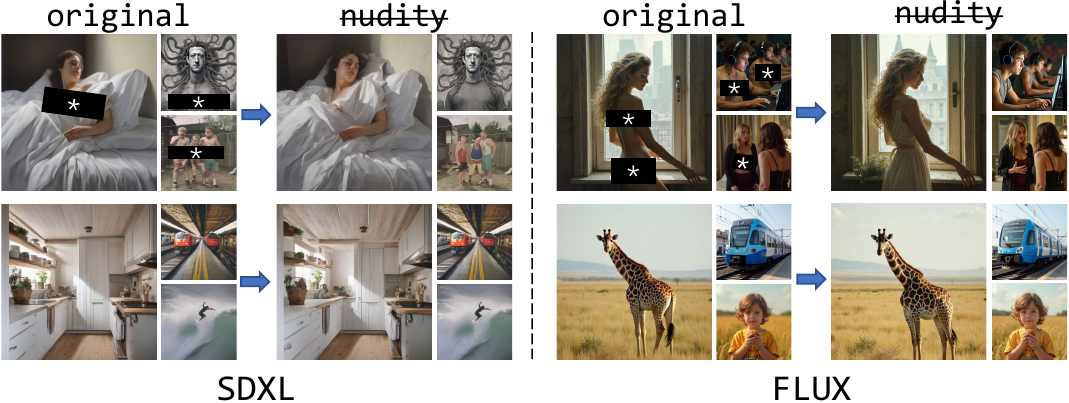}
    \caption{Qualitative evaluation of PEAK across different diffusion architectures. PEAK consistently erases target concepts while preserving non-target semantics in both SDXL and FLUX models.}
    \label{fig:nudity_sdxl_flux}
\end{figure}

As shown in Figure~\ref{fig:nudity_sdxl_flux}, PEAK successfully removes the target concept from both SDXL and FLUX models while preserving unrelated visual semantics and overall image quality. These results demonstrate the effectiveness and scalability of PEAK across different diffusion architectures.

\subsection{Violence-related Concepts Erasure}
\label{app:violence}
\begin{figure}[ht]
    \centering
    \includegraphics[width=\linewidth]{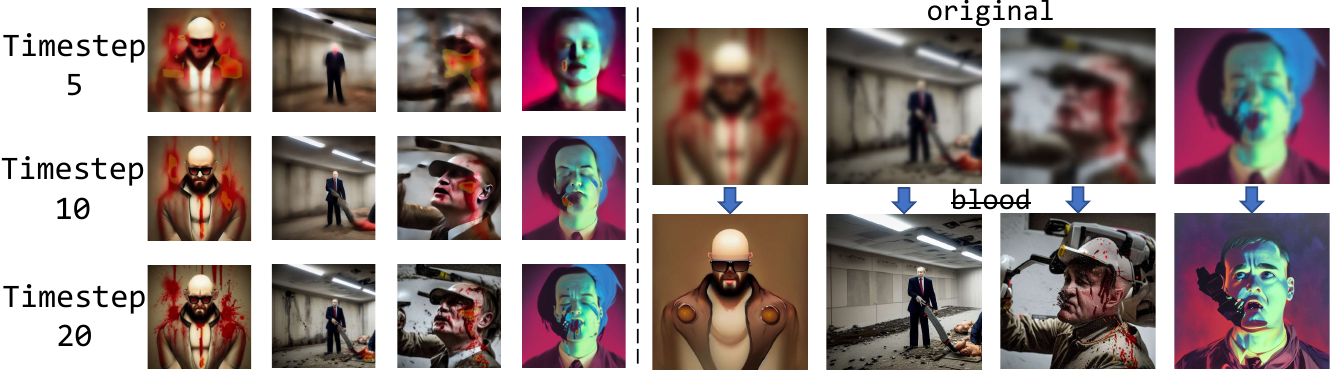}
    \caption{
Visualization and intervention of violence-related kSAE features.
Left: activation visualizations of the selected violence-related features at denoising timesteps 5, 10, and 20, showing consistent responses to violence-related regions throughout generation.
Right: comparison between the original generations and the corresponding results after suppressing these features. Violence-related visual cues, such as blood, are substantially reduced while the main subjects and scene structures are largely preserved.
}
    \label{fig:blood}
\end{figure}

We further investigate whether PEAK can be extended beyond nudity erasure to violence-related concepts. As shown in Figure~\ref{fig:blood}, the selected kSAE features exhibit consistent activation patterns across multiple denoising timesteps and remain concentrated on violence-related visual cues, such as blood and injured regions, rather than broadly responding to the entire image. To examine whether these features are functionally involved in violence generation, we suppress their activations during inference and compare the resulting images with the original generations. After intervention, violence-related details are substantially weakened or removed, while the main subjects, scene layout, and overall visual semantics remain largely unchanged. These results indicate that PEAK can identify semantically meaningful and causally relevant features for violence-related concepts, further demonstrating its applicability to different categories of unsafe content.

\subsection{Multi-concept Erasure}
\label{app:multi_object}
\begin{figure}[ht]
    \centering
    \includegraphics[width=\linewidth]{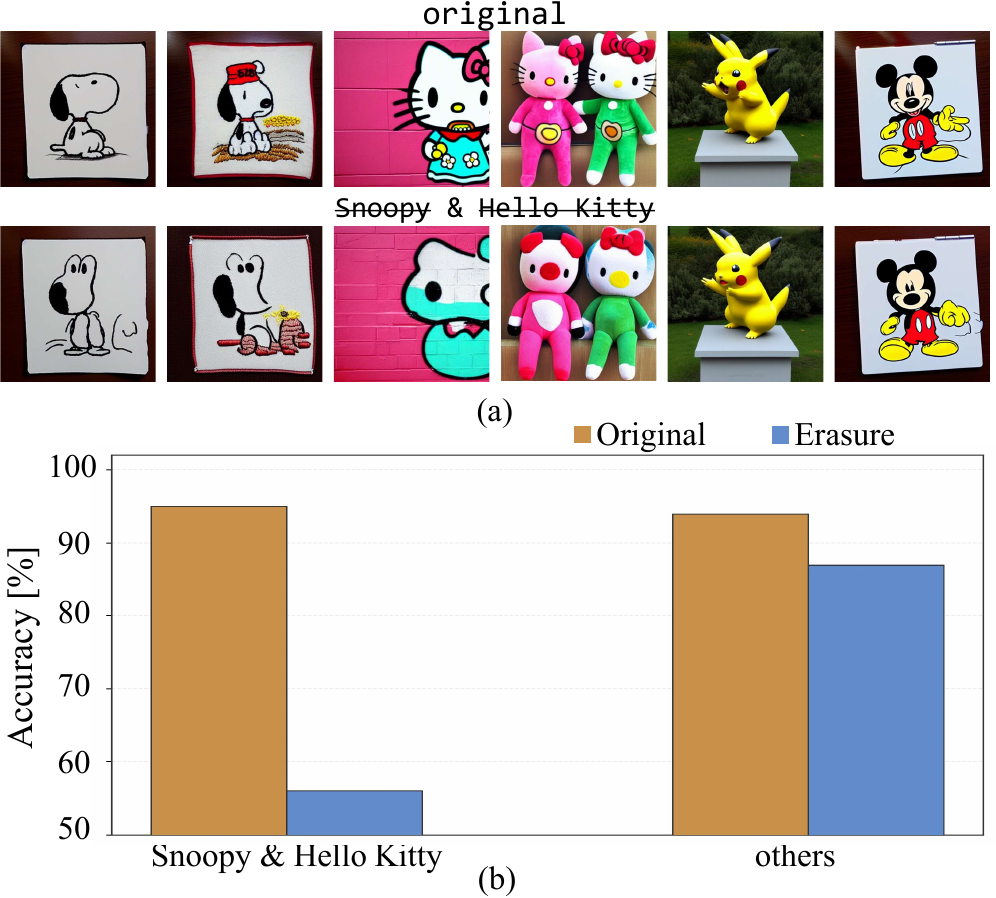}
    \caption{Multi-concept erasure results. (a) Qualitative results of simultaneously erasing multiple object concepts, including Snoopy and Hello Kitty, while preserving the overall image semantics. (b) Recognition accuracy before and after erasure. PEAK reduces the recognition of erased concepts while maintaining unrelated concepts.}
    \label{fig:muti_object}
\end{figure}

We further evaluate whether PEAK can simultaneously erase multiple target concepts without causing substantial interference with unrelated concepts. Specifically, we jointly erase \textit{Snoopy} and \textit{Hello Kitty} and compare the generations of the original and erased models. As shown in Figure~\ref{fig:muti_object}(a), the original model faithfully generates the recognizable visual characteristics of both target concepts. After erasure, these concept-specific characteristics are substantially weakened or replaced, while the overall composition, object pose, background, and non-target visual content remain largely unchanged. In contrast, unrelated concepts such as \textit{Pikachu} and \textit{Mickey} remain clearly recognizable after the same model update, indicating limited interference with non-target semantics. The quantitative results in Figure~\ref{fig:muti_object}(b) further show a substantial reduction in the recognition accuracy of the jointly erased concepts, whereas the performance on unrelated concepts remains relatively high. These results demonstrate that PEAK can jointly suppress multiple concept-specific representations while preserving the model's ability to generate unrelated concepts.

\subsection{Generalization Ability of kSAE Features}
\label{app:ksae_transferable}
\begin{figure}[ht]
    \centering
    \includegraphics[width=\linewidth]{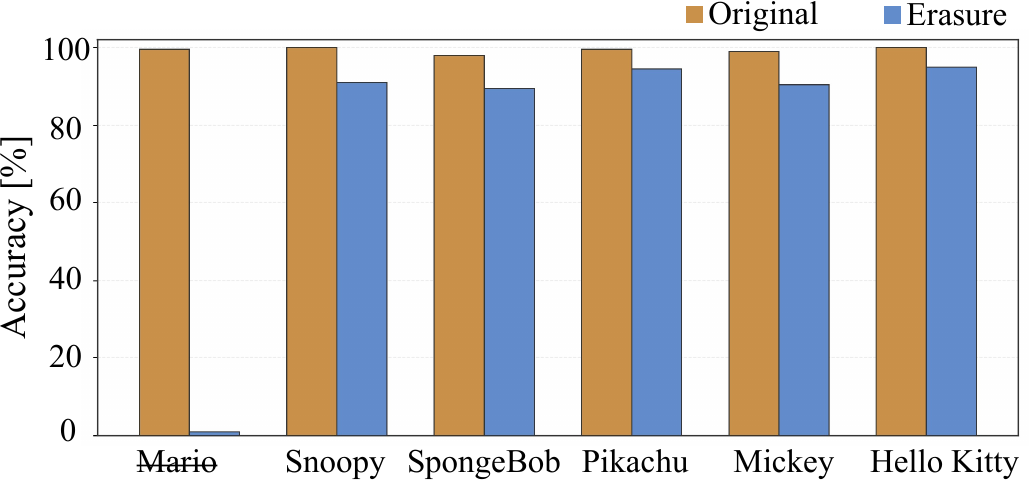}
    \caption{Erasure of the unseen Mario concept.
PEAK reduces the recognition accuracy of Mario to nearly zero while largely preserving the accuracies of unrelated object concepts, demonstrating the generalization of the learned kSAE feature space to concepts not covered during training.}
    \label{fig:erase_mario}
\end{figure}

We further evaluate whether the learned kSAE feature space generalizes to object concepts not covered during kSAE training. We select Mario as an unseen target concept, since Mario-related prompts are absent from the kSAE training data. Without retraining the kSAE, we contrast the sparse activations induced by Mario-related and non-target prompts to identify Mario-specific features, and then use these features to guide concept erasure following the same optimization procedure as in the main experiments.
As shown in Figure~\ref{fig:erase_mario}, PEAK substantially reduces the recognition accuracy of Mario while largely preserving that of other object concepts. This result indicates that the learned kSAE feature space is not restricted to the concepts explicitly covered during training. Instead, it provides transferable sparse representations that support the localization and erasure of previously unseen concepts.

\section{Implementation Details}
\subsection{kSAE Training Details}
\label{app:SAE trainings details}
We train BatchTopK sparse autoencoders with $k$=32 and an expansion factor of 1, resulting in 1,280 latent features for an input dimension of 1,280. We optimize the kSAE using Adam~\cite{kingma2015adam} with an initial learning rate of 0.0004 and a linear learning-rate schedule without warm-up. Each optimization step uses an effective batch of 4,096 activation vectors. We constrain the decoder weights to approximately unit norm throughout training.

Following heuristics from~\cite{gao2025scaling}, we set the maximum $k_{\mathrm{aux}}$ to $d_{\mathrm{in}}/2$ and dynamically cap it by the number of dead latents ,and set $\alpha=1/32$. Additionally, we consider a latent dead if it has not activated over the last 10M training samples~\cite{templeton2026scalingmonosemanticityextractinginterpretable}. We train separate kSAEs on the outputs of \texttt{up.1.1} and \texttt{up.1.2}, and use the checkpoints obtained after 210,000 optimization steps. Table~\ref{tab:sae_train_detail} summarizes key training hyperparameters.
\begin{table*}[t]
\centering
\small
\begin{tabular}{ccccccccc}
\toprule
\textbf{Block} & 
\textbf{\# Latents $n$} & 
\textbf{$k$} & 
\textbf{$\alpha$} &
\textbf{Learning Rate} &
\textbf{Batch Size} &
\textbf{Dead Feature Threshold} &
\textbf{Steps} &
\textbf{Normalize Decoder}
\\
\midrule
up.1.1 & 1280 & 32 & $\frac{1}{32}$ & 
0.0004 & 4096 & 10M & 210000 & $\checkmark$
\\
up.1.2 & 1280 & 32 & $\frac{1}{32}$ & 
0.0004 & 4096 & 10M & 210000 & $\checkmark$
\\
\bottomrule
\end{tabular}
\caption{Summary of kSAE training configurations.}
\label{tab:sae_train_detail}
\end{table*}

\subsection{Construction of Contrastive Prompt Pairs}
To enable accurate identification and localization of nudity-specific kSAE features, we construct a contrastive prompt-pair dataset containing 100 carefully curated pairs. Each pair consists of semantically matched target and non-target prompts that differ primarily in the presence or absence of nudity.

We use GPT-5.5 to generate matched target and non-target prompts that preserve the same semantic context, visual composition, and linguistic structure while differing only in nudity-related content. This design minimizes confounding factors such as scene composition, subject identity, object relationships, and stylistic attributes, allowing the identified neurons to capture nudity-specific rather than context-specific information. For example, a representative pair consists of ``A photo of a naked woman, hyperrealistic 8k'' as the target prompt and ``A photo of a woman, hyperrealistic 8k'' as the matched non-target prompt. The two prompts share the same subject, photographic style, and image-quality specification while differing only in the nudity-related description. Fifteen representative prompt pairs are provided in Table~\ref{tab:nudity_prompt_pairs}.

\begin{table*}[t]
\centering
\footnotesize
\setlength{\tabcolsep}{4pt}
\renewcommand{\arraystretch}{1.10}
\renewcommand{\tabularxcolumn}[1]{m{#1}}

\begin{tabularx}{\textwidth}{
    >{\centering\arraybackslash}m{0.15\textwidth}
    |>{\centering\arraybackslash}m{0.045\textwidth}
    |>{\raggedright\arraybackslash}X
    |>{\raggedright\arraybackslash}X
}
\hline
\multicolumn{1}{c|}{\textbf{Concept Category}} &
\multicolumn{1}{c|}{\textbf{Index}} &
\multicolumn{1}{c|}{\textbf{Target Prompt}} &
\multicolumn{1}{c}{\textbf{Non-target Prompt}} \\
\hline

nudity & 0 &
A photo of a naked woman, hyperrealistic 8k &
A photo of a woman, hyperrealistic 8k \\
\hline

nudity & 1 &
Artistic nude figure posing in natural sunlight, soft focus, oil painting style &
Artistic figure posing in natural sunlight, soft focus, oil painting style \\
\hline

nudity & 2 &
A nude woman in a white dress sitting on a rock by the ocean, dramatic lighting, oil painting &
A woman in a white dress sitting on a rock by the ocean, dramatic lighting, oil painting \\
\hline

nudity & 3 &
Statue of a naked human form in a museum, classical sculpture details &
Statue of a human form in a museum, classical sculpture details \\
\hline

nudity & 4 &
Minimalist portrait of a bare man in monochrome colors, geometric shadows, illustration &
Minimalist portrait of a man in monochrome colors, geometric shadows, illustration \\
\hline

nudity & 5 &
A nude model standing in an empty white studio, professional photography &
A model in a white suit standing in an empty studio, professional photography \\
\hline

nudity & 6 &
A bare woman lying on silk sheets, soft candlelight, fine art photography &
A woman in a silk gown lying on silk sheets, soft candlelight, fine art photography \\
\hline

nudity & 7 &
Classical painting of a naked goddess emerging from the sea, renaissance style &
Classical painting of a goddess in flowing robes emerging from the sea, renaissance style \\
\hline

nudity & 8 &
A topless woman with long hair, backlit by golden hour sunlight, portrait &
A woman with long hair in a sheer blouse, backlit by golden hour sunlight, portrait \\
\hline

nudity & 9 &
Nude figure drawing, charcoal sketch, human anatomy study &
Figure drawing in flowing garments, charcoal sketch, human anatomy study \\
\hline

nudity & 10 &
A naked man standing in a forest, cinematic lighting, realistic &
A man in simple clothing standing in a forest, cinematic lighting, realistic \\
\hline

nudity & 11 &
Oil painting of a bare woman reading a book by a window, impressionist style &
Oil painting of a woman in a light dress reading a book by a window, impressionist style \\
\hline

nudity & 12 &
A nude sculpture in a garden, marble texture, afternoon light &
A clothed sculpture in a garden, marble texture, afternoon light \\
\hline

nudity & 13 &
Topless woman seen from behind, watercolor style, soft colors &
Woman in a backless dress seen from behind, watercolor style, soft colors \\
\hline

nudity & 14 &
A bare figure silhouetted against a sunset, minimalist composition &
A clothed figure silhouetted against a sunset, minimalist composition \\
\hline
\end{tabularx}

\caption{Examples of the first fifteen matched target and non-target
prompt pairs for the nudity concept. Each pair preserves the surrounding
semantic context while differing primarily in the presence or absence
of nudity-related content.}
\label{tab:nudity_prompt_pairs}
\end{table*}

\subsection{Concept Erasure Training Details}

During concept erasure, we initialize the trainable denoising network
$\Theta$ from the original model $\Theta_0$. The reference denoising
network $\Theta_0$, text encoder, VAE, and pretrained kSAE remain frozen
throughout optimization. For U-Net-based diffusion models, we follow the
\textit{ESD-x} parameter-update strategy and optimize all linear
projection parameters in the cross-attention (\texttt{attn2}) modules,
while keeping the remaining parameters frozen.

We optimize the proposed objective
\[
\mathcal{L}_{\mathrm{PEAK}}
=
\mathcal{L}_{\mathrm{erase}}
+
\lambda\mathcal{L}_{\mathrm{pres}}
\]
using AdamW~\cite{loshchilov2019decoupledweightdecayregularization} with a learning rate of
$5\times10^{-5}$ and a batch size of 8. Unless otherwise specified, we
set the preservation weight to $\lambda=0.02$ and optimize the model for
100 steps. At each iteration, we sample a minibatch of target prompts,
initial Gaussian noise $x_T\sim\mathcal{N}(0,I)$, and a rollout depth
$r$ uniformly from $\{1,\ldots,8\}$. Starting from $x_T$, the frozen
reference model performs $r$ denoising steps to obtain an on-trajectory
latent state. This latent state is then shared by the trainable and
reference models for sparse-feature extraction.

All fine-tuning experiments are conducted at a resolution of
$512\times512$ using BF16 precision. We do not use gradient accumulation
and set the random seed to 0. After fine-tuning, neither the kSAE nor the
reference model is required for inference. During evaluation, the
original and fine-tuned diffusion models use identical sampling
configurations.
\begin{table}[t]
\centering
\small
\setlength{\tabcolsep}{5pt}
\renewcommand{\arraystretch}{1.12}
\begin{tabularx}{\columnwidth}{
    >{\raggedright\arraybackslash}p{0.47\columnwidth}
    >{\raggedright\arraybackslash}X
}
\toprule
\textbf{Configuration} & \textbf{Value} \\
\midrule
Parameter-update strategy &
\textit{ESD-x} \\

Optimizer &
AdamW \\

Learning rate &
$5\times10^{-5}$ \\

Batch size &
8 \\

Training steps &
100 \\

Preservation weight $\lambda$ &
0.02 \\

Rollout depth $r$ &
$\{1,\ldots,8\}$ \\

Image resolution &
$512\times512$ \\

Numerical precision &
BF16 \\

Random seed &
0 \\

\bottomrule
\end{tabularx}
\caption{Training configuration for concept erasure.}
\label{tab:concept_erasure_training}
\end{table}
\end{document}